# AliMe KG: Domain Knowledge Graph Construction and Application in E-commerce

Feng-Lin Li
fenglin.lfl@alibaba-inc.com
Alibaba Group

Hehong Chen
hehong.chh@alibaba-inc.com
Alibaba Group

Guohai Xu
guohai.xgh@alibaba-inc.com
Alibaba Group

Tian Qiu
qiutian.qt@alibaba-inc.com
Alibaba Group

Feng Ji
zhongxiu.jf@alibaba-inc.com
Alibaba Group

Ji Zhang
zj122146@alibaba-inc.com
Alibaba Group

Haiqing Chen
haiqing.chenhq@alibaba-inc.com
Alibaba Group

## ABSTRACT

Pre-sales customer service is of importance to E-commerce platforms as it contributes to optimizing customers' buying process. To better serve users, we propose **AliMe KG**, a domain knowledge graph in E-commerce that captures user problems, points of interests (POI), item information and relations thereof. It helps to understand user needs, answer pre-sales questions and generate explanation texts. We applied AliMe KG to several online business scenarios such as shopping guide, question answering over properties and recommendation reason generation, and gained positive results. In the paper, we systematically introduce how we construct domain knowledge graph from free text, and demonstrate its business value with several applications. Our experience shows that mining structured knowledge from free text in vertical domain is practicable, and can be of substantial value in industrial settings.

## CCS CONCEPTS

• **Computing methodologies** → **Information extraction**; **Semantic networks**; • **Information systems** → **E-commerce infrastructure**.

## KEYWORDS

E-commerce, Pre-Sales Customer Service, Domain Knowledge Graph

## 1 INTRODUCTION

Understanding customers' buying process is of key importance to E-commerce platforms as it directly contributes to improving Conversion Rate (CVR). For years, E-commerce sites such as Amazon, Alibaba and Jingdong have been employing search box and catalog as the primary way for customer interaction, teaching customers to search items through product-oriented keywords, and helping them to quickly find what they want to buy. To bridge the semantic gap between what users want in their mind and how items are organized in E-commerce platforms, Luo et al. [10] have proposed to capture user needs in product knowledge graph and associate them with items. Although there have been many efforts to improve search engine and recommender system [3, 17–19, 25], there are still room left for optimizing customers' buying process.

The main reason, in our observation, is that E-commerce sites usually assume users know exactly what they need and offer search engine as a tool in support of customer interaction. However, in many cases, what users have in mind is their problems (e.g., "dry skin"), and they do not have a clear idea about the solution (i.e., their true needs, e.g., "preserve moisture"), which need to be inferred based on domain knowledge. Also, customers often need to seek more information outside E-commerce sites when assessing whether a product can truly meet their needs or resolve their problems. That is, customers need a service, rather than merely a tool, when making a purchasing decision. The service should be able to talk with customers, infer their needs, provide userful information about specific items, give suggestions and reasonable explanations. This is where pre-sales customer service comes in.

With the fast development of Deep Learning (DL) techniques, chatbot has become a natural choice for E-commerce customer service as it is able to automatically answer customers' questions and largely improve the efficiency of customer support service. In light of this trend, we have launched AliMe [7] for a real-world industrial application that offers pre-sales customer service for hundreds of thousands of stores on the Alibaba E-commerce platform[1]. To better understand users, we constructed **AliMe KG**, a domain knowledge graph in the field of E-commerce that captures user problems, points of interest (POI), item information, and relations thereof. The KG serves as the foundation for recognizing user problems (e.g., "dry skin 皮肤干"), inferring user needs (e.g., "preserve moisture 保湿"), completing item information (e.g., "hyaluronic acid 玻尿酸" *cause* "preserve moisture 保湿"), and generating explanatory recommendation reasons (e.g., "we recommend this facial cleanser as it contains hyaluronic acid and is able to preserve moisture, which perfectly fits your dry skin problem").

---
[1] AliMe offers not only pre-sales retail service but also after-sales customer service. We focus on the pre-sales part in this paper.

In this paper, we present AliMe KG, introduce our systematic construction methodology and semi-automated knowledge mining process, and demonstrate its business value through several realistic applications in pre-sales conversation scenarios.

Our paper makes the following contributions:

- We propose AliMe KG, an ongoing domain knowledge graph that currently supports the top-50 main categories on the Alibaba E-commerce platform, and enables our chatbot to understand user problems, infer user needs, answer pre-sales questions and provide explanation texts.
- We present a systematic methodology and semi-automated processes for mining structured knowledge from free texts, and introduce our innovations in the underlying key components, namely Phrase Mining, Named Entity Recognition and Relation Extraction.
- We applied AliMe KG to several realistic applications and demonstrated its business value through online A/B tests.

The rest of the paper is organized as follows: Sec. 2 introduces our motivation; Sec. 3 presents an overview of AliMe KG; Sec. 4 systematically describes how the KG is constructed; Sec. 5 shows the evaluation results of main building blocks; Sec. 6 demonstrates its application and business value; Sec. 7 reviews related work; Sec. 8 discusses future work and concludes the paper.

## 2 MOTIVATION

In general, customers' buying process includes five stages[2]: need recognition, information search, evaluation of alternatives, purchase decision and post-purchase behavior. Over the years, search engine and recommender system are the principal means that E-commerce platforms used to enhance user shopping experience. Customers have been accustomed to finding out solutions to their problems or needs through general search engines (e.g., Baidu) or vertical websites (e.g., Zhihu), and then searching specific items at online shopping sites. On observing the semantic gap between user needs and product taxonomy, Luo et al. [10] have proposed to capture user needs as E-commerce concepts (e.g., "outdoor barbecue") and then associate the concepts with product items (e.g., grills and butter). In this way, E-commerce search engines are able to understand user concepts and accordingly recommend associated items. Despite the efforts, search engine by itself is insufficient to further optimize customers' buying process as it is limited to part of the buying process – namely the information search stage – in customers' mental model.

Pre-sales customer service is a more natural way for understanding users through conversational interaction. Nowadays chatbot has been prevailing in E-commerce customer service, either pre-sales or after-sales, since it is able to reduce large amount of time spent on customer enquiries. We have already launched AliMe [7] for a real-world industrial application. Although it is able to achieve high performance through text classification and/or matching over question answer (QA) pairs, it is still hard to say our chatbot can "understand" customer questions.

To better serve customers, our chatbot need to recognize user problems, infer user needs, provide useful item information, and generate explanation texts based on domain knowledge. Therefore, we propsoed AliMe KG, a domain knowledge graph in E-commerce, and applied it to empower AliMe for a better user experience.

## 3 KG OVERVIEW

We show our core ontology in Fig. 1. Three commonly accepted concepts, namely "User", "Item" and "Scenario", are adopted from classic buying process: a user intent to buy some items at/for a certain scenario. Note that "Scenario" refers to not only shopping place (e.g., city, shop) but also various kinds of consumption scenarios (e.g., Teachers' Day, outdoor barbecue). The concept "IPV" captures property values of items (Item-Property-Value, e.g., "cleansing foam 洁面泡沫"-*ingredient*-"bisabolol 红没药醇"). Two new concepts, **Problem** and **POI**, are our key contributions. "Problem" refers to a problematic state that a user is at (e.g., "pimple 长痘痘"), "POI" captures users' need or solution to user problem ("anti-acne 清痘抑痘"). Also, there are two types of newly added links: *need*, which relates problem to POI (e.g., "pimple" *need* "anti-acne"), and *cause*, which links IPV and POI (e.g., "bisabolol" *cause* "anti-acne"). These links are established based on domain knowledge.

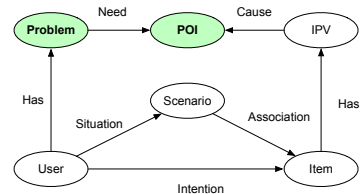

**Figure 1: The core ontology of AliMe KG**

In general, our KG consists of three layers: User, POI and Item. As shown in Fig 2, the user layer captures users' problems (e.g., "pimple 长痘痘"). The item layer records items' properties (e.g., "ingredient 成分") and property values (e.g., "bisabolol 红没药醇") based on their categories. The POI layer, which serves as the bridge between users and items, on one hand is used to link users' problem (e.g., "pimple" *need* "anti-acne"), on the other hand is used to relate items' property value (e.g., "bisabolol" *cause* "anti-acne"). That is, the *need* link can be used to infer users' need based on their problems, and the *cause* link can be used to retrieve corresponding items and explain why a product has a feature of concern.

There are three points to be noted. First, items in Alibaba (also other E-commerce sites) are organized based on Category-Property-Value (CPV): thousands of categories form a hierarchical structure and leaf categories have properties pre-defiend, items are associated with leaf categoris and accordingly instantiate their predefined properties. Second, our KG captures domain knowledge at class level, not instance level. For example, user concern "pimple 长痘痘" and product feature "anti-acne 清痘抑痘" mentioned above are class level concepts. When put into use, they will be instantiated by specific users and items. Third, dotted links are added or predicted through KG completion. For instance, if a user concern about "pimple", which needs "anti-acne", we will add a *preference* link between the user and "anti-acne".

---
[2]https://en.wikipedia.org/wiki/Buyer_decision_process

Table 1: Types, formats and sources of domain knowledge

| Type | Format | Example | Source |
| --- | --- | --- | --- |
| User POI | Category - Has - POI | "clothing" - *has_poi* - "skin-friendly" | item articles, detail pages |
| User Probelm | User - Has - Problem | "user" - *has_problem* - "pimple" | chatlog |
| CPV&IPV | Category - Property - Value<br>Item - Property - Value | "cleansing foam" - *ingredient* - "bisabolol"<br>"cleansing foam" - *ingredient* - "bisabolol" | item articles, detail pages |
| POI Knowledge | Problem - Need - POI<br>IPV - Cause - POI | "pimple" - *needs* - "anti-acne"<br>"bisabolol" - *cause* - "anti-acne" | chatlog |

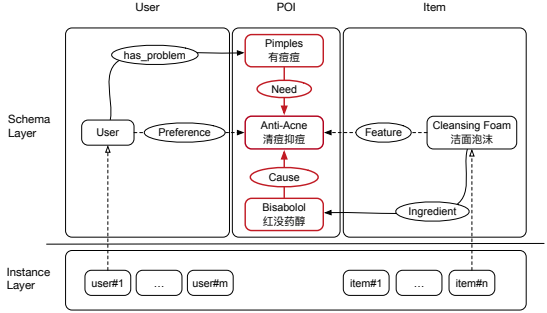

Figure 2: An excerpt of our domain knowledge graph

## 4 KG CONSTRUCTION

In this section, we first introduce the types of knowledge in our KG, then describe our methodology about how to acquire knowledge.

As shown in Table 1, our KG includes the following types of knowledge: user problems, POIs, CPV&IPV data, and POI relational knowledge. Among them, user problems and POIs are captured in phrases. We classify a phrase as a user problem if it describes a problematic state that users want to find out solutions for (e.g., "pimple 长痘痘"), and as a user POI if it reveals user interests or potential needs (e.g, "anti-acne 清痘抑痘"). CPV and IPV data are mainly imported from the Alibaba product knowledge graph and complemented according to our application in pre-sales conversation scenarios. POI knowledge is relational and encodes the associations of POIs with user problems, and with CPV&IPV data.

We present our knowledge mining process in Fig. 3. In general, it includes two parts: *node mining* and *link prediction*. The process takes as input data source, which includes chatlog, item detail pages and item articles, firstly extract nodes, then establish links, and finally output structured knowledge. During the process, crowdsourcing is employed as the primary way for KG quality inspection. Each sub-process, including crowdsourcing[3], has been automated in our production environment and is scheduled to run periodically.

---
[3]We have integrated the Alibaba crowdsourcing platform into our knowledge mining process, as such we are able to automatically deploy tasks and recycle labelled data.

### 4.1 POI Mining

The goal of POI mining is to extract potential user interests or needs. For example, "skin-friendly 亲肤", "anti-acne 清痘抑痘" and "safe and non-toxic 安全无毒", are typical POIs in Clothing, Beauty and Tableware. As users are often not aware of or do not explicitly express their POIs, we choose E-commerce content (item detail pages and articles), rather than customer service chatlog, as our data source for POI mining.

Given a domain (i.e., first-level categories), we first retrieve the set of leaf categories and associated items from product knowledge graph, and then obtain detail information and articles for each item from the Alibaba E-commerce content platform. Subsequently, as shown in Fig. 3a, we use heuristic rules and phrase mining to obtain phrases, from which we collect both positive and negative samples through crowdsourcing annotation. After that, we train a binary BERT [4] classifier to predict whether an extracted phrase is a POI. The obtained POI candidates will be further checked by crowd annotators, and finally those accepted will be added into our knowledge graph and organized as "Category - *has* - POI" tuples.

*4.1.1 Phrase Mining.* Phrase mining is of importance to both POI mining and user problem mining (to be detailed in Sec. 4.2). We adopt the automated phrase mining approach [15], extend it with deep semantic features for quality phrase classification, and further employ BERT masked language model (MLM) for pruning.

We describe the algorithm flow in Fig. 4. The process takes as input a corpus from which phrases to be extracted and a lexicon consists of more than one million words accumulated in practice, and output a set of quality phrases. Specifically, we start collecting phrase seeds without human labor: the procedure extracts text sequences that are separated by punctuations and within certain length as raw phrases, and treats the intersection of frequent raw phrases and lexicon words as phrase seeds. Next, we establish the set of phrase candidates that contains all the n-grams over a certain threshold (e.g., 3) in the corpus, and label those candidates as positive if they are in the seed pool and negative otherwise. Often, the number of positives is rather small (e.g., hundreds) while that of negatives is large (e.g, hundreds of thousands). As suggested by AutoPhrase [15], we train a random forest (RF) classifier by drawing $K$ (e.g., 100) phrase candidates with replacement from the positives and negatives respectively for each base decision tree classifier.

For feature design, except statistic features such as frequency, tf-idf, pointwise mutual information[4] and information content[5],

---
[4]https://en.wikipedia.org/wiki/Pointwise_mutual_information
[5]https://en.wikipedia.org/wiki/Information_content

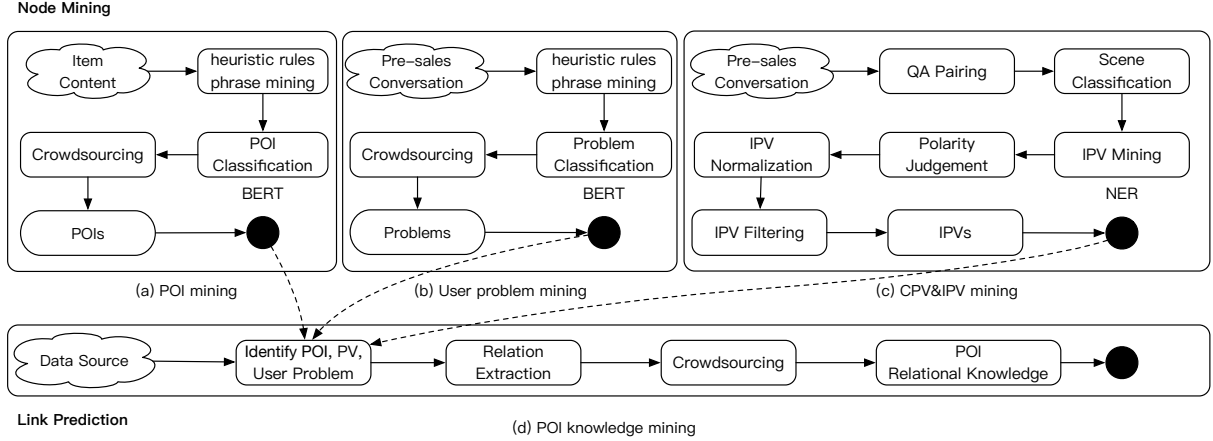

Figure 3: Knowledge mining process

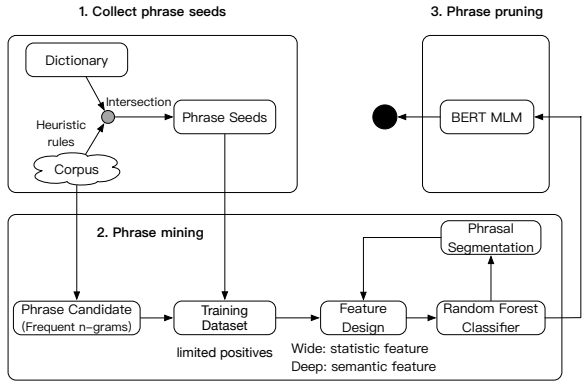

Figure 4: Overview of phrase mining model

we also consider semantic features that average character embeddings. As some of the n-grams are invalid phrases and the text sequences that they located in need to be properly segmented, we further use POS-guided phrasal segmentation [15] to rectify the frequency of phrase candidates. At last, we adopt BERT MLM to filter out those incomplete ones. Specifically, we mask the first (resp. last) token of a phrase candidate, and feed it into BERT to check if the masked token is in the top-$N$ (e.g., 1) predicted token list.

*4.1.2 Binary Classifier.* Once we have collected training samples, we train a binary BERT classifier for subsequent POI mining. Specifically, we continually train BERT on E-commerce corpus, and construct samples by concatenating a phrase $p$ and its leaf category $lc$ in the form of "[CLS] $p$ [SEP] $lc$ [SEP]" for both training and prediction, where "[CLS]" and "[SEP]" are conventional tokens adopted from BERT, and leaf category $lc$ is treated as the context of $p$.

## 4.2 User Problem Mining

We treat user problems as problematic states that users are in and extract them from pre-sales conversation between customers and service staff/chatbot. As shown in Fig. 3b, which is similar to POI mining, we extract candidate phrases through heuristic rules and phrase mining, collect training samples through crowdsourcing, and train a binary BERT classifier for user problems classification. One difference, however, is that training and serving samples are in the form of "[CLS] $p$ [SEP] $d_p$ [SEP]", where $p$ indicates a phrase, and $d_p$ refers to one of the sentences that $p$ is located in. Note that user problems in our KG are captured as conceptual knowledge and associated with the "User" concept, and are employed for problem-recognition in online conversations for particular users.

## 4.3 CPV & IPV Mining

Although we are able to import all items from the Alibaba knowledge graph, CPV&IPV data can be missing. Therefore, we also need to complement missing properties and property values of concern. For simplicity, we fix the properties and enumerable property values of each leaf category, and focus on mining missing property values. For example, the color of T-shirt can be red, blue, white and so on. Supposing that the color of a particular T-shirt is missing, we try to fill the gap with one of the enumerated values through IPV mining. If the missing value of a property is not enumerable, we only accept it if it exceeds a certain frequency in our data source.

The IPV mining process is shown in Fig. 3c. We start with pre-sales conversations in a specific domain, collect QA pairs about items, filter out pairs irrelevant to business scenarios, then extract property values through NER. After that, we judge the polarity to check whether an extracted value belongs to a particular item, normalize property value based on synonyms, and at last use predefined CPV dictionary for quality control.

*4.3.1 Named Entity Recognition (NER).* In our case, NER is used to identify property values from a text sequence and then assign them with property labels. For example, in the question "Can it be used by pregnant women? 孕妇能用吗", "pregnant women 孕妇" is the value of property "target users 适用人群". Understanding text from such property-value perspective is foundational in E-commerce as it can be used not only for offline knowledge mining but also for online question answering over properties.

Adopting the unified Embedder-Encoder-Decoder framework [8], we extend the standard BiLSTM-CRF model [5] with BERT. In our proposed model (Fig. 5), BERT extracts features and provides contextual embeddings for input tokens, BiLSTM acts as an encoder, and CRF predicts their final labels. Besides, we incorporate external lexicon (words only) and dictionary (words with type, e.g., words can be property values and type can be properties in our case) knowledge through introducing extra features.

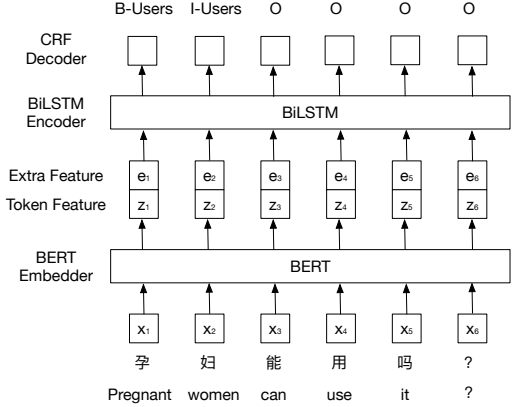

**Figure 5: Overview of BERT-BiLSTM-CRF NER model**

Given an input sequence of tokens $\{x_1, x_2, x_3, ..., x_m\}$, we obtain the representation of $x_i$ ($1 \leq i \leq m$) through concatenating contextual feature $z_i$ from BERT and extra feature $e_i$ constructed based on lexicon/dictionary, as formulated in Equation 1.

$$x_i = z_i \oplus e_i \quad (1)$$

For extra features, we first employ a word segmenter with external lexicon to segment each input sequence and accordingly obtain a sequence of softwords [13, 24], and then use BMES scheme, rather than BIO, to represent the positional information of each token in a softword. For example, for the token "pregnant 孕" in the softword "pregnant women 孕妇", we give it a label "B" that represent the begining of a softword and assign it a $d$-dimentional (e.g., 256) randomly initialized embedding. For external dict, we also add word type information to each token in a softword. For instance, if we know "pregnant women 孕妇" is of type "target users", then we will label the token "pregnant 孕" as "B#target_users", which will also be associated with a different embedding.

*4.3.2 Polarity Judgement.* Polarity judgement is a must in IPV mining becasue negative polarity would introduce incorrect info. For example, in the sentence "The T-shrt is not red", the extracted value "red" is not the color of the particular T-shirt. We currently use heuristic rules for polarity judgement, and will change to DL models in the near future.

## 4.4 POI Relational Knowledge Mining

To support user need inference and explain while an item has a feature of concern, we need to relate POI with user problems and IPVs, respectively. The key idea of link prediction between two concepts is to see whether we are able to find one or more text sentences that reveal a specific kind of relation between them.

We summarize POI relational knowledge mining in Fig.3d. Taking the *cause* relation for example, we first collect text sentences from E-commerce content, and then identify CPVs (resp. POIs) for each sentence though NER (resp. dictionary match). We keep those sentences that include both CPV (e.g., "bisabolol 红没药醇") and POI (e.g., "anti-acne 清痘抑痘"), and let crowd annotators check whether a sentence reveal the desired relation (e.g., *cause*) between them. At last, we recycle labeled data, and train an classification model for subsequent link prediction. The mining of *need* triples is similar, but with the data source extended to Baike corpus.

Note that links are established between POIs and CPVs, and instantiated to IPVs through inheritance when put into use. For example, conceptually we have "bisabolol 红没药醇" *cause* "anti-acne 清痘抑痘", where "bisabolol" is the value of property "ingredient" of a specific category, if an item at that category has "'bisabolol" as its ingredient, the item would also possess that feature.

*4.4.1 Relation Extraction.* Relation extraction is used to predict whether two concepts form a specific relation in a certain context (i.e., the sentence they are located in). As in [20], we base our model on BERT and incorporate information from anchor concepts. For a sentence $s$ with two anchor concepts $c_1$ and $c_2$, we insert a special token '$' (resp. '#') at the begining and end of $c_1$ (resp. $c_2$) in the input sentence, and use the "CLS" token embedding together with concept embeddings for final classification (see Fig. 6).

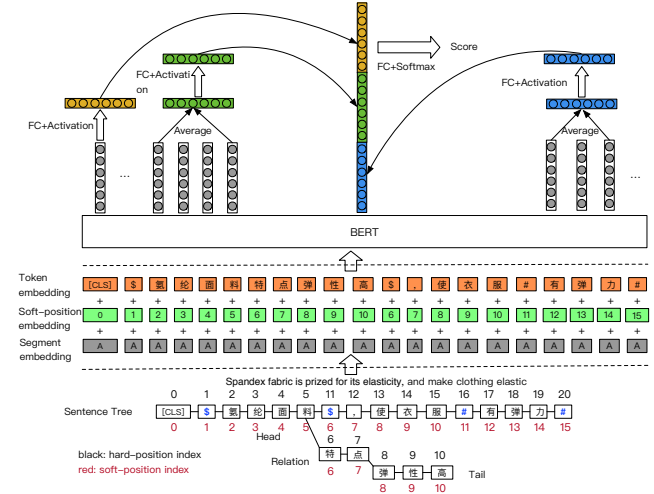

**Figure 6: Overview of relation extraction model**

We also borrow the idea of triple knowledge injection from K-BERT [9] to incorporate external knowledge (e.g., HOWNET [14], CNDBpedia [22]). Specifically, we use anchor concepts $c_1$ and $c_2$ to query triples $E$ that with them as heads from a knowledge graph $\mathcal{K}$, and inject the triples into the sentence to form a new sentence $s_{new}$ through soft position. As in Fig 6, there are two concepts "spandex fabric 氨纶面料" and "elastic 有弹力" in the given text. The first concept is associated with a triple "spandex fabric 氨纶面料" - *characteristic* - "high elasticity 弹性高", which is injected into the original sentence through soft position (red indices). We formualte this process as Equation 2.

$$E = query(\{c_1, c_2\}, \mathcal{K}) \qquad (2)$$
$$s_{new} = inject(s, E)$$

*4.4.2 Illustrative Example.* We use an illustrative example to demonstarte our POI knowledge mining process. As shown in Fig. 7, given the sentence "Food grade silicone does not contain BPA, is resistant to high temperature sterilization, hence it is a truly reassuring and safe tableware", we first identify "Food grade silicone" as a type of "material" in the Baby Tableware category, and recognize "high temperature sterilization", "reassuring" and "safe" as POIs. After that, we establish a *cause* link between the CPV and each POI, capture them as triples as shown in the bottom right corner of Fig. 7

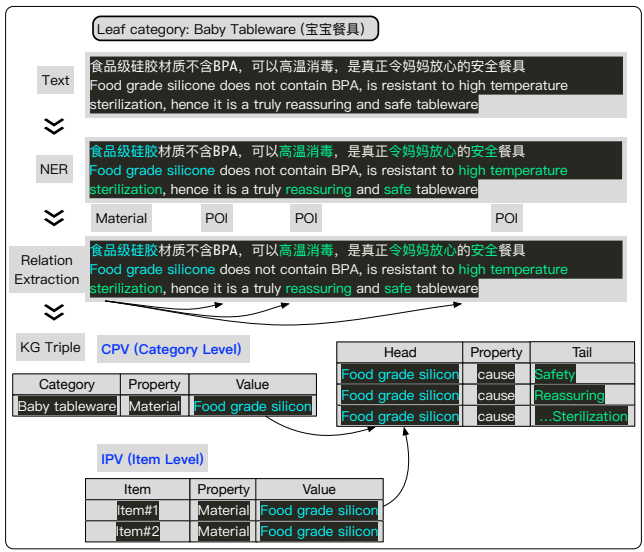

Figure 7: Example of POI relational knowledge mining

## 5 EVALUATION

In this section, we first provide a statistical overview of our KG, and then present detailed evaluations for our key components. Note that we use precision $p$ to evaluate phrase mining (whether an extracted text sequence is a meaningful/quality phrase), $F_1$ to measure NER, and $AUC$ to measure binary classification tasks.

### 5.1 KG Statistics

We show the statistics of AliMe KG in Table 2. So far we have accumulated 365K$^+$ POIs, 1K$^+$ user prblems and 29K$^+$ CPVs, 8.6K$^+$ "User problem - POI" triples, and 113K$^+$ C-PV-POI triples. We constrcuted an active item pool consists of 600K$^+$ items, and collected 3,300K$^+$ IPV triples, 43.95% of which are acquired through our IPV mining. Further, by applying C-PV-POI triples to items, we obtained more than 13,590K$^+$ I-PV-POI triples via inheritance.

### 5.2 Evaluations

*5.2.1 KG Data Quality.* Conceptual level knowledge, including POIs, user problems, CPVs, the assocaitions of POIs with user problems, and with CPVs, has been completely checked by crowdsourcing, hence its quality can be ensured. Instance level knowledge, IPV data and I-PV-POI triples, was only spot-checked because of its huge amount. Our spot-check also shows the accuracy of IPV and I-PV-POI triples is of high quality and can be directly used in real-world applications.

Table 2: The statistics of AliMe KG

| Layer | Knowledge | Numbers | Source | Qulity Control |
|---|---|---|---|---|
| User Layer | User problem | 1K$^+$ | Mining (100%) | Complete |
| Item Layer | CPV | 29K$^+$ | Import (100%) | Complete |
| | Item | 600K$^+$ | Import (100%) | - |
| | IPV | 3,300K$^+$ | Import (56.05%) Mining (43.95%) | Spot |
| POI Layer | POI | 365K$^+$ | Mining (100%) | Complete |
| | User problem - POI | 8.6K$^+$ | Mining (100%) | Complete |
| | C - PV - POI | 113K$^+$ | Mining (100%) | Complete |
| | I - PV - POI | 13,590K$^+$ | Mining (100%) | Spot |

*5.2.2 Phrase Mining.* We evaluated our phrase mining method with an E-commerce corpus consists of 830K$^+$ text sentences. We randomly selected 200 phrase candidates obtained from each method and assess whether a candidate is a quality phrase through crowd-sourcing. We show the results in Table 3, where "Unsupervised" model represents the one use left&right entropy and PMI for phrase candidate scoring, "RF" indicates random forest, "SEG" means POS-guided phrasal segmentation. One can see that the phrasal segmentation and BERT MLM are very helpful in improving the precision of phrase candidates. When combined together, our model is able to achieve a precision of 88% without any manual annotation, which is practically usable in industrial settings.

Table 3: Experimental results of phrase mining

| Model | #Phrase Candidate | $p$ |
|---|---|---|
| Unsupervised | 28716 | 58.7% |
| RF | 35566 | 48.5% |
| RF+SEG | 13516 | 70% |
| RF+MLM | 12984 | 80% |
| RF+SEG+MLM | 7514 | 88% |

Our model is able to support million-scale corpus. Moreover, it is domain independent and does not need human annotation, which are favored by the industry. Besides, as we can also see that the number of extracted quality phrases decreases as precision increases, we will further improve our model to achieve a better trade-off.

*5.2.3 Named Entity Recognition.* We evaluated our NER model in the Beauty domain with 6 property types. We show the results in Table 4. We can see that the incorporation of lexicon knowledge is able to achieve an increase of 1.56% in $F_1$, and that of dict knowledge brings an increase of 3.14%.

*5.2.4 Relation Extraction.* We assessed our relation extraction model on *cause* link prediction in the Clothing domain. As in Table 5, although baseline model is able to achive high performance on this binary classification task ($AUC = 95.12\%$), the incorporation of external knowledge still further improves the result: an increase of 0.28% for HOWNET and 0.24% for CNDBpedia in *AUC*.

**Table 4: Experimental results of named entity recognition**

| Model | $F_1$ |
|---|---|
| BERT+BiLSTM+CRF | 78.17% |
| +Lexicon | 79.73% |
| +Dict | 81.31% |

**Table 5: Experimental results of relation extraction**

| Model | AUC |
|---|---|
| BERT + concept info (Baseline) | 95.12% |
| +HOWNET | 95.40% |
| +CNDBpedia | 95.36% |

## 6 KG APPLICATIONS

AliMe KG has been applied to several real-world applications in pre-sales conversation scenarios, including but not limited to shopping guide, question answering over properties, and explanatory recommendation reason generation. We show how AliMe KG contributes to optimizing customers' buying process in Table 6, and introduce the applications in the following sections.

**Table 6: How AliMe KG optimizes customer buying process**

| Ends | Means | Application |
|---|---|---|
| Need recognition | Infer user needs | Shopping guide |
| Information search | - | - |
| Evaluation of alternatives | Provide product info | Question answering over properties |
| Purchase decision | Provide explanations | Recommendation reason generation |
| Post-purchase behavior | - | - |

### 6.1 Shopping Guide

AliMe [7] is a real-world chatbot application that offers pre-sales custromer service for hundreds of thousands of stores on the Alibaba platform. Being similar to offline shopping in physical stores, customers often describe their problems and ask for suggestions or inquire about a particular product. For example, "My skin is a bit dry, what kind of facial cleanser is suitable? 我的皮肤有点干，适合什么洗面奶". If we directly use the extracted keywords "skin is a bit dry" and "facial cleanser" to query search engines, the results are often somehow irrelevant.

To answer such knowledge-oriented questions, we applied AliMe KG for *query rewriting* and *item recall*. On one hand, the search keywords of the aforementioned query will be re-written as "preserve moisture 保湿" according to the domain knowledge "dry skin 皮肤干" - *need* - "preserve moisture 保湿" captured in our KG, and then used to query the Alibaba product graph. On the other hand, we maintain an active item pool ourselves and construct inverted index for items in the form of "$POI - item\#1, item\#2...$" based on I-PV-POI knowledge aforehand for item recall in subsequent recommendation. Our A/B test in the Beauty&Personal Care domain shows that our KG is able to cover 5% of the pre-sales conversations and bring a relative increase of $20\%^+$ in Conversion Rate (CVR).

### 6.2 Question Answering over Product Property

Customers in pre-sales conversations would also ask detailed product questions to seek more information for making purchasing decisions. For example, "Can it be used by pregnant women?" queries about the property "target users". For this purpose, we employed KBQA (question answering over knowledge base) [6] to answer questions concerning product properties. Specifically, we use NER to recognize properties and/or preoperty values from questions, judge customers' intention (which property is being queried if multiple ones are mentioned), and then retrieve corresponding property value as answer. Compared with previous approach that refer customers to product pages for detailed product questions, we are able to provide accurate answers with AliMe KG.

### 6.3 Recommendation Reason Generation

When customers query about or when we recommend a specific item, we can also provide a recommendation reason to explain why the product is suitable based on our domain knowledge graph. For example, when customers send an "sweater 卫衣" item link in AliMe, we can retrieve from our KG its style "round neck 圆领" and associated POIs "cute 可爱" and "leisure 休闲", then we are able to generate an explanatory recommendation reason "This sweater has a cute round neck, and brings a feeling of cute and leisure 这件卫衣的领子是圆形的款式，显得非常的可爱和休闲" using graph-to-sequence generation technique [23]. We have applied our KG for generating explanatory recommendation reasons mainly in the Clothing domain. Online tests show that our KG is able to cover 30% of the pre-sales conversations and the CVR gains a relative increase of 4.8%.

## 7 RELATED WORK

There have been many efforts on establishing open domain KG such as WordNET [11], Freebase [2] and DBpedia [1]. Unlike early lexical knowledge bases such as WordNET [11] and HowNET [14] that are mainly established manually by experts, our KG is constructed semi-automatically, with crowdsourcing involved in the process. Being different from Freebase [2] and DBpedia [1] that focus on describing facts with well defined types, we lay emphasis on the schema layer, where the POI concepts and relations thereof are extracted from natural language text.

NELL [12] tries to automatically extract triples from web with an initial ontology defining categories (e.g., Athlete, Sport) and binary relations (AthletePlaySport), but has a limited precision and scale of concepts. Probase [21] provides a large-scale probabilistic *IsA* concept taxonomy in support of text understanding. Further, ConceptNet [16] inlcudes common sense knowledge by capturing informal relations between concepts, which are words or phrases of natural language that conceptualize general human knowledge (e.g., "hot weather" *CauseDesire* "turn on air conditioner"). Our KG captures conceptual level knowledge (e.g., "dry skin" *need* "preserve moisture"), which is more akin to ConceptNet, but in a vertical domain.

In the field of E-commerce, Amazon has proposed "Product Knowledge Graph (PG)" with the puprose of answering any question about products and related questions, but currently do not focus on user needs. The most similar to us is AliCoCo [10], which enrichs the product taxnomy with E-commerce concepts that capture user needs, and accordingly associates the new concepts with items for search and recommendation. We differ from their work in two aspects. First, we have different purposes and applications: AliMe KG is mainly used for pre-sales conversation while AliCoCo is mainly designed for search and recommendation. Second, more importantly, our KG captures domain knowledge which enables to infer user needs and explain why a product is fit for certain user problems, which contributes to optimizing customers' buying process as we have discussed in Sec. 6.

## 8 DISCUSSION AND CONCLUSION

### 8.1 Lessons Learned

In this paper, we demonstrated that constructing industrial scale KG from free text in vertical domain is practicable, and showed how to reduce manual annotation and utilize external knowledge to improve model performance. Our approach also has limitations on quality control as concept level knowledge need to be fully checked at the moment, and need to be further improved.

There are three points to be noted when building a knowledge graph. First of all, set a clear objective. There are some questions to ask before diving into technique details: what are the (incremental) value of KG? What are the application scenarios of KG? Is deep learning (e.g., text classification and matching) sufficient? Second, design a feasible technological path. For example, keep schema simple as complex schema brings poor scalability, redude human annotation costs as you can. Third, do not forget data quality control as industrial applications often impose strict quality requirements on data and knowledge, which are often not easy to satisfy.

### 8.2 Conclusion

In this paper, we propose AliMe KG, a domain knowledge graph designed for better connecting customers to items through pre-sales chatbot conversation. We systematically introduce how it is constructed from free text semi-automatically, and demonstrate how it contributes to optimizing customers' buying process through several real-world applications.

On one hand, we will continually enlarge AliMe KG to cover the majority of vertical domains on the Alibaba E-commerce platform. On the other hand, there is a new trend of influencer marketing and live streaming in E-commerce, for which we need multi-modality item content, including but not limited to text, image and video. Multimodal KG is our key topic for the next step.